\pdfoutput=1

\documentclass[11pt]{article}

\PassOptionsToPackage{backref=page}{hyperref}
\usepackage[]{emnlp2023}
\usepackage{times}
\usepackage{latexsym}

\usepackage[T1]{fontenc}

\usepackage[utf8]{inputenc}

\usepackage{microtype}

\usepackage{times}
\usepackage{latexsym}

\usepackage{url}
\usepackage{amsmath}
\usepackage[nameinlink]{cleveref}
\crefformat{section}{\S#2#1#3} 
\crefformat{subsection}{\S#2#1#3}
\crefformat{subsubsection}{\S#2#1#3}
\crefformat{paragraph}{\S#2#1#3}
\crefname{figure}{Figure}{Figures}
\crefname{table}{Table}{Tables}
\usepackage{enumitem}
\usepackage[colorinlistoftodos]{todonotes}
\usepackage{tabularx}
\usepackage{booktabs}
\usepackage{multirow}
\usepackage{multicol}
\usepackage{fixltx2e}
\usepackage[most]{tcolorbox}
\usepackage{adjustbox}
\usepackage{booktabs}
\usepackage[font=small,labelfont=bf]{caption}
\usepackage{subcaption}
\usepackage{soul}
\usepackage{float}
\usetikzlibrary{shapes}

\usepackage{scalerel,xparse}
\NewDocumentCommand\githubicon{}{\includegraphics[scale=0.025]{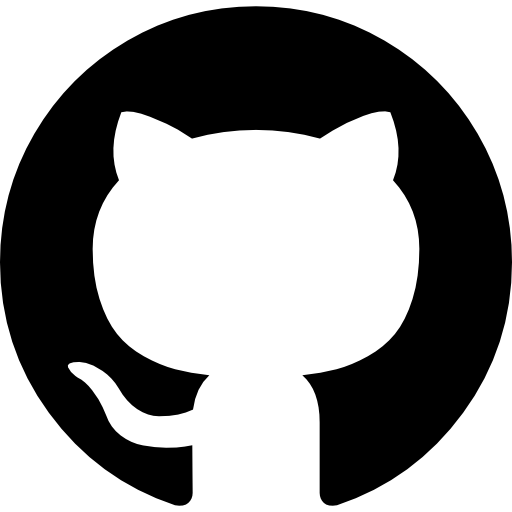}}

\usepackage{longtable}
\usepackage{makecell}

\newcommand{\roberta}{\texttt{RoBERTa}}
\newcommand{\bert}{\texttt{BERT}}

\newcommand{\pos}{\text{Part of Speech}}
\newcommand{\ner}{\text{Named Entity Recognition}}

\newcommand{\coref}{\text{Coreference Resolution}}

\newcommand{\spr}{\text{Semantic Proto-Role Labeling}}

\newcommand{\dpr}{\text{Definite Pronoun Resolution}}

\newcommand{\ontonotes}{\text{OntoNotes}}
\newcommand{\ontonotespos}{\text{OntoNotes-PoS}}
\newcommand{\ewtpos}{\text{EWT-PoS}}
\newcommand{\ptbpos}{\text{PTB-PoS}}
\newcommand{\ewtsyndepcls}{\text{EWT-Syn-Dep-Cls}}
\newcommand{\ewtsyndeppred}{\text{EWT-Syn-Dep-Pred}}
\newcommand{\ptbsyndepcls}{\text{PTB-Syn-Dep-Cls}}
\newcommand{\ptbsyndeppred}{\text{PTB-Syn-Dep-Pred}}
\newcommand{\ontonotessrl}{\text{OntoNotes-SRL}}
\newcommand{\ontonotesconst}{\text{OntoNotes-Const}}
\newcommand{\ontonotesner}{\text{OntoNotes-NER}}
\newcommand{\conll}{\text{CoNLL}}
\newcommand{\conllner}{\text{CoNLL-2003-NER}}
\newcommand{\conllchunking}{\text{CoNLL-Chunking}}
\newcommand{\sprone}{\text{SPR-1}}
\newcommand{\sprtwo}{\text{SPR-2}}
\newcommand{\dprd}{\text{DPR}}
\newcommand{\semeval}{\text{Semeval-Rel-Cls}}
\newcommand{\ontonotescoref}{\text{OntoNotes-Coref}}

\usepackage{mdframed}
\AtBeginEnvironment{blockquote}{\small}
\definecolor{block-gray}{gray}{0.85}
\newtcolorbox{blockquote}{colback=block-gray,boxrule=0pt,boxsep=0pt,breakable}

\AtBeginEnvironment{definition}{\small}

\title{Implications of Annotation Artifacts in Edge Probing Test Datasets}

\author{
\parbox{\linewidth}{\centering
Sagnik Ray Choudhury$^{1*}$,
Jushaan Kalra$^{2*}$}\\ 
\parbox{\linewidth}{\centering
  $^{1}$University of Michigan,
  $^{2}$Wadhwani AI\\
  \texttt{sagnikrayc@gmail.com, jushaan18@gmail.com}
  }
  }
  
\date{}

\begin{document}
\maketitle

\begingroup\def\thefootnote{*}\footnotetext{The authors contributed equally.}\endgroup 

\begin{abstract}
Edge probing tests are classification tasks that test for grammatical knowledge encoded in token representations coming from contextual encoders such as large language models (LLMs). Many LLM encoders have shown high performance in EP tests, leading to conjectures about their ability to encode linguistic knowledge. However, a large body of research claims that the tests necessarily do not measure the LLM's capacity to encode knowledge, but rather reflect the classifiers' ability to learn the problem.
Much of this criticism stems from the fact that often the classifiers have very similar accuracy when an LLM vs a random encoder is used.
Consequently, several modifications to the tests have been suggested, including information theoretic probes.
We show that commonly used edge probing test datasets have various biases including memorization. When these biases are removed, the LLM encoders do show a significant difference from the random ones, even with the simple non-information theoretic probes \footnote{\githubicon\ The code is available at \url{https://github.com/Josh1108/EPtest.git}}.
\end{abstract}

\section{Introduction}
\label{sec:intro}

Word embeddings generated from large corpora can be expected to encode knowledge about syntax and semantics \cite{manning2020emergent}. This is certainly truer for the contextual ones from large language models such as \texttt{Elmo} \cite{peters-etal-2018-deep}, \bert\ \cite{DBLP:conf/naacl/DevlinCLT19} or \roberta \cite{DBLP:journals/corr/abs-1907-11692}. Edge probing (EP) tests \cite{liu-etal-2019-linguistic, tenney-etal-2019-bert} are standard classification tasks to probe for such knowledge.

Consider the sentence ``The Met is closing soon'', the word ``Met'' functions as a noun, referring to a museum rather than the past form of the verb ``meet''. To determine its part of speech, humans rely on the context words ``the'' and ``is''. If a classifier predicts this token as a noun using \textit{only} the representation from a contextual LLM encoder such as \bert\, (i.e., without using the entire sentence), it is implied that these contextual signals are encoded within the token representation itself. EP tests aim to uncover such syntactic and semantic knowledge encoded (\cref{sec:edge-probing}).

EP tests are however \textit{indirect} measures of such knowledge. A high accuracy of an encoder in an EP test for a grammatical property in itself does not necessarily guarantee that the said knowledge is encoded.
Instead, the score should be \textit{significantly higher} than the same from a baseline, which is typically set as static embedding encoders \cite{DBLP:journals/corr/abs-1709-04482} or contextual encoders with random weights \cite{zhang-bowman-2018-language, tenney-etal-2019-bert, liu-etal-2019-linguistic}. 

NLP tasks are typically modeled by datasets, albeit imperfectly  \cite{DBLP:conf/eacl/RavichanderBH21}, and consequently, the performance of the encoders in the EP tests are confounded by the choice of the test dataset and its inherent biases. Despite a long history of research in edge probing tests, this problem has not been studied well \cite{belinkov-2022-probing}. 

To bridge this research gap, we propose three research questions.

\textbf{RQ1: Are there ``annotation artifacts'' in the EP test datasets?} 
Many standard NLP datasets have data points that can be solved by superficial cues, i.e., reasoning strategies unrelated to the expected causal mechanism of the task at hand \cite{DBLP:conf/iclr/KaushikHL20}. For example, \citet{DBLP:conf/naacl/GururanganSLSBS18} show that a negation operator in the premise is a strong predictor of the ``contradiction'' class in the SNLI \cite{bowman-etal-2015-large} dataset. \citet{DBLP:conf/emnlp/SenS20} show that in popular extractive machine reading comprehension (MRC) datasets such as SQuAD \cite{DBLP:conf/emnlp/RajpurkarZLL16} or HotpotQA \cite{DBLP:conf/emnlp/Yang0ZBCSM18}, in many cases the answer phrase can be found in the first sentence of the context. We analyze 17 EP test datasets across 10 tasks and find different biases in multiple of them.

\textbf{RQ2: Do the EP models use heuristics?} Existence of annotation artifacts in the data does not necessarily imply that the models will learn to use the related heuristics, eg., predict ``contradiction'' whenever the premise contains a negation. We can a) remove the biased test data \cite{mccoy-etal-2019-right} or b) adversarially perturb it \cite{jia-liang-2017-adversarial} and observe the performance degradation (if any) of a model. A significant degradation will indicate that the model does depend on the heuristic. Using this technique, we show that the EP classifiers trained with random encoders do indeed learn to use the heuristics to a large extent, whereas the same ones trained with pre-trained encoders do not in the same capacity.

\textbf{RQ3: Do the pre-trained encoders encode grammatical knowledge better than the random encoders?} A strong criticism of EP tests is that often the performances of the pre-trained and the random encoders are not \textbf{significantly different} \cite{zhang-bowman-2018-language}. This is often attributed to the ``classifier knowledge'' problem, i.e., the EP classifier learns the task itself and does not necessarily depend on the encoder representations. Various information theoretic probes \cite{pimentel-etal-2020-information} have been proposed to solve this, including a popular one based on the Minimum Description Length (MDL) principle \cite{grunwald2000model}. In this MDL probe \cite{DBLP:conf/emnlp/VoitaT20}, a combined measure defined on the EP classifier model complexity and its performance is minimized. The MDL codelengths of contextualized representations such as \texttt{Elmo} are shown to be much lower than the corresponding random ones even when their EP test accuracies are very similar. However, we show this is not strictly necessary, and the similar performance of a pre-trained and random encoder can largely be attributed to the EP test dataset biases, as in when the ``biased'' data points are removed, a simple linear or MLP classifier shows a significant difference in the pre-trained vs random encoder.  We investigate this further and show that Bayesian classifiers such as MDL probes are not ``inherently better'' in testing an encoder's ability to encode grammatical knowledge. 

\section{Edge Probing}
\label{sec:edge-probing}

\subsection{Formulation}
\begin{figure}[!t]
  \centering
        \includegraphics[scale=0.6]{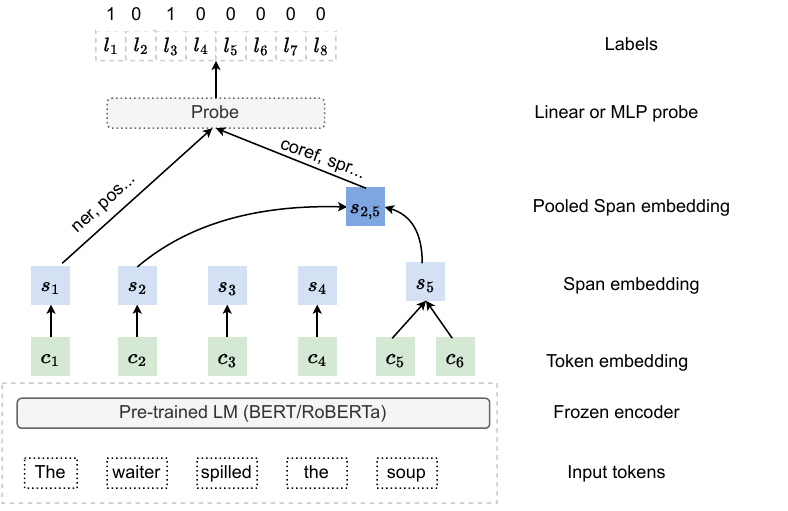}
    \caption{The architecture for edge probing tasks. } 
    \label{fig:ep-arch}
\end{figure}

We base our experiments on the model architecture (\cref{fig:ep-arch}) and edge probing tasks proposed by \citet{tenney-etal-2019-bert} and \citet{liu-etal-2019-linguistic}, two cotemporaneous works that introduced the idea of EP tests on contextual encoders.

Given a sentence $S = [T_{1},...T_{n}]$ of $n$ tokens, a span $s_{k} = [T_{i},...T_{j}]$ is defined as a contiguous sequence of tokens $i$ to $j$. Depending on the task, an individual or a pair of spans is assigned a label. For example, in the Named Entity Recognition EP test, the label of the span ``Barack Obama'' would be PERSON. In the EP test for Coreference Resolution, a pair of spans would be labeled true or false depending on whether they were co-referent to each other in a sentence or not. 

The input to the EP classifier is an embedding $e_i \in \mathcal{R}^d$ for a (pair of) span(s) and its goal is to predict its label. Token representations can be generated from the top layer \cite{DBLP:conf/iclr/TenneyXCWPMKDBD19} or the intermediate layers \cite{liu-etal-2019-linguistic} of an encoder, which is typically a large language model (LLM) such as \bert, \roberta, or \texttt{Elmo}. For our EP tests, we consider the top-layer representations.\footnote{\citet{DBLP:conf/iclr/TenneyXCWPMKDBD19} uses both the top layer and a mixed representation from all layers, and \citet{hewitt-manning-2019-structural} uses the top layer. As there is not a significant difference in the mixed vs top layer representations in \citet{DBLP:conf/iclr/TenneyXCWPMKDBD19}, we leave the mixed representations for future work.} Following \citet{liu-etal-2019-linguistic}, we generate $e_i$ by taking an average of all token embeddings in the span, which is further averaged over the spans in the two-span tasks.

 The final embedding is passed to an EP classifier (also referred to as a probe), which is either a) \textbf{MLP}: A multilayer perceptron with \textbf{one} hidden layer ($1024$ dim) and a RELU activation, or b) \textbf{Linear}: A linear layer without any non-linearity. For all models, the dropout \cite{10.5555/2627435.2670313} is kept at \texttt{1e-1}. 
 
\citet{liu-etal-2019-linguistic} used a linear layer classifier, and so did \citet{DBLP:conf/iclr/TenneyXCWPMKDBD19}, who also used a single hidden layer MLP. Follow-up work by \citet{DBLP:conf/emnlp/HewittL19} and \citet{DBLP:conf/emnlp/VoitaT20} both used single or multiple hidden layer MLPs, but we didn't find much difference in our experiments by increasing the number of layers. 
 Specifically, \citet{DBLP:conf/emnlp/HewittL19} suggested using probes with high ``selectivity'', i.e., they should have a high accuracy on an EP task, but a low score when the labels of the same task are randomized (control tasks). They concluded that simpler, i.e., lower depth probes showed higher selectivity, which is another reason for our probe choice. 
 
 Crucially, during training, \textit{only the parameters of the probe are changed} and the encoder below is kept frozen. If the LLM encoder truly encodes certain types of syntactic (eg., identifying constituent types) or semantic (coreference relation between phrases) knowledge, we can expect it to have a significantly higher performance in the related EP test than an encoder of the same architecture but with random weights.

\subsection{Edge Probing Tasks and Datasets}

To ensure wide coverage, we experiment with 17 EP datasets involving 10 different NLP tasks that have been used before in \citet{DBLP:conf/iclr/TenneyXCWPMKDBD19} and \citet{liu-etal-2019-linguistic}. The tasks are described below, the dataset statistics are presented in \cref{table:dataset}. 

\begin{table}[!t]
\footnotesize
\centering
\begin{tabular}{  p{\dimexpr 0.45\linewidth-2\tabcolsep} 
 p{\dimexpr 0.25\linewidth-2\tabcolsep} 
 p{\dimexpr 0.15\linewidth-2\tabcolsep} 
 p{\dimexpr 0.15\linewidth-2\tabcolsep} }
\toprule
Dataset & \multicolumn{3}{l}{\#Points in the EP test data} \\
& Train    & Test  & Dev  \\ \cmidrule(l){1-4}
\multicolumn{4}{c}{\textbf{{\pos\ Tagging}}} \\
\ewtpos\textsuperscript{2}      & $204,607$  & $25,097$        & $25,150$\\
\ptbpos\textsuperscript{2}      & $950,028$  & $56,684$        & $40,117$\\
\ontonotespos\textsuperscript{1}   & $2,070,382$        & $212121$        & $290,013$\\
\multicolumn{4}{c}{\textbf{\ner}} \\
\conllner\textsuperscript{2}       & $203,621$  & $46,435$        & $51,362$\\
\ontonotesner\textsuperscript{1}   & $128,738$  & $12,586$        & $255,133$\\
\multicolumn{4}{c}{\textbf{\coref}} \\
\dprd\textsuperscript{1}   & $1,787$    & $949$   & $379$\\
\ontonotescoref\textsuperscript{1} & $207,830$  & $27,800$        & $26,333$\\
\multicolumn{4}{c}{\textbf{Syntactic Dependency Classification}} \\
\ewtsyndepcls\textsuperscript{2}   & $203,919$  & $25,049$        & $25,110$\\
\ptbsyndepcls\textsuperscript{2}   & $910,196$  & $54,268$        & $38,417$\\
\multicolumn{4}{c}{\textbf{Syntactic Dependency Prediction}} \\
\ewtsyndeppred\textsuperscript{1,2}  & $383,462$  & $45,901$        & $46,155$\\
\ptbsyndeppred\textsuperscript{2}  & $1,820,225$        & $108,529$       & $76,820$\\
\multicolumn{4}{c}{\textbf{Semantic Proto-Role Labeling}} \\
\sprone\textsuperscript{1} & $7,611$    & $1,055$ & $1,071$\\
\sprtwo\textsuperscript{1} & $4,925$    & $582$   & $630$\\
\multicolumn{4}{c}{\textbf{One Task Datasets}} \\
\conllchunking\textsuperscript{2}  & $211,727$  & $47,377$        & - \\
\ontonotesconst\textsuperscript{1} & $1,851,590$        & $190,535$       & $255,133$\\
\ontonotessrl\textsuperscript{1}   & $598,983$  & $61,716$        & $83,362$ \\
\semeval\textsuperscript{1}        & $8,000$    & $2,717$ & -     \\ \bottomrule
\end{tabular}
\caption{Statistics for the EP datasets used in this paper, with the tasks and in which paper they were used in: \citet{DBLP:conf/iclr/TenneyXCWPMKDBD19}\textsuperscript{1} or \citet{liu-etal-2019-linguistic}\textsuperscript{2}.}
\label{table:dataset}
\end{table}

\textbf{\pos\ Tagging.} POS tagging is a syntactic task, where each token is assigned one of the possible part-of-speech tags. e.g. ``$[Napoleon]_{NNP}$ Bonaparte was the emperor of France'', where NNP stands for ``Proper Noun, Singular''. We use 3 different datasets for this task: the \ontonotes\ corpus \cite{ontonotes-corpus}, the Penn
Treebank (PTB) corpus \cite{ptb-dataset} and the Universal Dependencies English Web Treebank (EWT) corpus \cite{ewt-dataset}. 

\textbf{\ner.} 
NER is a task to predict the pre-defined semantic category of a span such as persons, organizations, date, and quantity, e.g. - ``$[Napoleon \: Bonaparte]_{PERSON}$ was the emperor of France.'' We use the \ontonotes\ corpus and the \conll\ 2003 shared task dataset \cite{conll-2003-shared}. 

\textbf{Constituency Labeling.}
The goal of this task is to recover the constituency parse tree of a sentence, eg., ``$[Napoleon \: Bonaparte]_{NP}$ was the emperor of France.'', where NP stands for ``Noun Phrase''. We use the \ontonotes\ corpus for this task.

POS, NER, and Constituency Labeling are usually modeled as token-level \textit{tagging} tasks using the standard BIO format \cite{pradhan-etal-2013-towards} but in the EP tests, they are classification problems. The classifier predicts the label for a token or a span, which can be one of the pre-defined ones, eg., ``ADJ'' for \pos, ``PER'' for NER, or ``PP'' for Constituency Labeling or ``None'' if the input can not be assigned a label. Importantly, the classifier has access to only the token representations and not the whole sentence. 

\textbf{Coreference Resolution.}
Coreference resolution is the task of finding anaphoric relations between spans in a text:  e.g. ``$[Barack \: Obama]_1$ is an ex-US president, $[He]_2$ lives in DC with his wife Michelle.'' In the EP tests, this reduces to a binary classification task: given two spans, predict whether they refer to each other (``Barck Obama'', ``he'': true) or not (``Michelle'', ``he'': false). We use the \ontonotes\ corpus as well as the Definite Pronoun resolution (DPR) dataset \cite{dpr}, which is considered more challenging.

\textbf{Semantic Role Labeling.}
In the SRL task, the goal is to understand \textit{semantic} roles (who did what to whom and when) between spans (argument) in a sentence and a verb (predicate): eg., ``$[The \:waiter]_{AGENT}$ $[spilled]_{VERB}$ $[the \: soup]_{THEME}$. In the EP tests, this is modeled as a two-span multi-class classification task for which the \ontonotes\ corpus is used.

\textbf{Chunking.}
While a constituency parse of a sentence is a hierarchical structure, chunking \cite{abney1992parsing} divides the text into syntactically related non-overlapping groups of words. We use the \conll-2000-Chunking corpus \cite{tjong-kim-sang-buchholz-2000-introduction}. For the EP tests, this is a one-span multi-class classification problem.

\textbf{Semantic Proto-Role Labeling.}
Proposed by \citet{reisinger-etal-2015-semantic}, this is a task of annotating detailed, non-exclusive semantic attributes, such as change of state or awareness, over predicate-argument pairs as in SRL. Similar to the SRL EP test, this is modeled as a two-span classification problem, but as there can be more than one potential attribute of the predicate-argument relation, this is a multi-label task. We used two datasets, \sprone\ \cite{spr1-dataset}, and \sprtwo\ \cite{rudinger-etal-spr-2-datset}, derived from the Penn Treebank and the English Web Treebank respectively.

\textbf{Relation Classification.}
Initially proposed by \cite{rel-cls}, Relation Classification is the task of predicting the relation that holds between two nominals, from a given knowledge base. We use the SemEval dataset from \cite{hendrickx-etal-2010-semeval}. For the EP tests, this reduces to a two-span multi-class classification task.

\textbf{Syntactic Dependency Classification.}
Given representations of two tokens from a sentence, $[head]$ and $[mod]$, the task is to predict the syntactic relationship between the two.
We use the Penn Treebank \cite{ptb-dataset} and English Web Treebank \cite{ewt-dataset} datasets. For EP tests, this boils down to a two-span multi-class classification task.

\textbf{Syntactic Dependency Prediction.}
The goal of this task is to find whether a dependency arc exists between two tokens in their syntactic structure. We use the Penn Treebank and the English Web Treebank, the same as in the classification variant. This is a two-span binary classification task for EP tests.

Where development data was not available from the source, 10\% of the data from the training set was reserved for validation. In a few other cases, the testing set had labels not present in the training set, these data points were discarded. The final datasets (bar the licensed ones) will be made available.

\section{Annotation Artifacts in EP Test Datasets}
\label{sec:rq1}

Our analysis indicates that almost all EP test datasets have a significant repetition bias: many samples in the training data are repeated in the test. However, their labels may always not be the same, for example, in the NER EP test, the span ``Google'' might have the label ``ORG'' or ``O'' depending on whether the span refers to the company or the search engine developed by it. 

We ask two questions. In a test dataset, in what percentage of cases a test data point is in the training data and has only one label? For example, in the NER datasets, if the span ``Google'' appears in both the training and the test dataset with the \textit{only} label ``Org'', the EP classifier can successfully classify it by memorization. We call it the \textbf{Mem-Exact} heuristic.

Even if the training data contains multiple labels for a span (eg., both ``ORG'' and ``O''), the EP classifier might be able to successfully classify it in the test data by simply learning the label distribution for the span and not the inherent contextual relationships. In the \textbf{Mem-Freq} heuristic we find the percentage of test data points that are present in the training data and can be classified correctly using the training label distribution. We also consider a baseline: the \textbf{Mem-Uniform} heuristic where instead of the true label distribution the class labels can be predicted by sampling from a uniform distribution. 

\cref{tab:heuristic-result} shows that a large percentage of data points indeed can be classified heuristically, i.e., the dataset has significant biases. Importantly, if an EP classifier does adopt a heuristic, it would need no specific representation for the spans, let alone from a pre-trained or a random one.

\begin{table}[!t]
\footnotesize
\centering
\begin{tabular}{  p{\dimexpr 0.4\linewidth-2\tabcolsep} 
 p{\dimexpr 0.2\linewidth-2\tabcolsep} 
 p{\dimexpr 0.2\linewidth-2\tabcolsep} 
 p{\dimexpr 0.2\linewidth-2\tabcolsep} }
\toprule
Dataset & Mem-Exact  & Mem-Freq  & Mem-Uniform \\ \cmidrule(l){1-4}
\ewtpos      & $89.73$  & $42.85$        & $48.03$\\
\ptbpos      & $97.11$  & $42.03$        & $52.62$\\
\ontonotespos   & $98.06$        & $65.40$      & $35.26$\\
\conllner       & $86.87$  & $28.15$        & $67.93$\\
\ontonotesner   & $70.53$  & $23.40$        & $55.58$\\
\dprd   & $28.98$    & $0.21$   & $15.81$\\
\ontonotescoref & $36.55$  & $16.64$        & $26.51$\\
\ewtsyndepcls   & $37.98$  & $4.56$        & $34.30$\\
\ptbsyndepcls   & $62.17$  & $12.50$        & $51.81$\\
\ewtsyndeppred  & $42.44$  & $13.38$        & $31.99$\\
\ptbsyndeppred  & $68.04$ & $17.37$       & $47.75$\\
\sprone & $5.2$    & $0.47$ & $4.55$\\
\sprtwo & $7.2$    & $0.42$   & $1.37$\\
\conllchunking  & $89.89$  & $57.72$        & $33.88$\\
\ontonotesconst & $45$        & $17.79$       & $33.57$\\
\ontonotessrl   & $32.07$  & $6.98$        & $26.76$ \\
\semeval        & $3.35$    & $0.11$ & $3.3$     \\ \bottomrule
\end{tabular}
\caption{Accuracy (in \%) of the heuristic algorithms.}
\label{tab:heuristic-result}
\end{table}

\section{Do the EP Models Use Heuristics?}  
\label{sec:rq2}
Based on the dataset biases discovered in \cref{sec:rq1}, we hypothesize that the EP classifiers can use heuristic algorithms, but there will be a difference in the random vs pre-trained encoders. Specifically, \textit{EP test classifiers with random encoders will learn to use various heuristics} as the input representations themselves do not provide much information. On the other hand, the same classifier models with pre-trained encoders will tend to not make use of such heuristic mechanisms. If the hypothesis is true, we will see a \textit{significant drop in the performance with the random encoders compared to the pre-trained encoders} when the ``heuristically classifiable'' data points are removed from the test data. 

\subsection{Experimental Setup}
We use 4 encoders - \bert\ (the base-cased version), \roberta\ (the base version), and their randomized versions. Following \citet{DBLP:conf/iclr/TenneyXCWPMKDBD19}, the random encoders are the same LLM models randomly initialized \cite{xavier_initial} as it is done before pre-training.

For each encoder and EP classifier model (Linear and MLP, see \cref{sec:edge-probing}) we train 3 models.\footnote{Each model was trained for 3 epochs with a batch size of 16 using the AdamW optimizer \cite{DBLP:journals/corr/KingmaB14}, a learning rate of \texttt{1e-3} and a linear warmup learning rate scheduler \cite{howard-ruder-2018-universal}.} The models showed little variance on the test data (within 0.1\% of the average), therefore, we chose the best model for the subsequent experiments. 

\subsection{Results and Analysis}

\begin{table*}[!hbt]
\footnotesize
\centering
\begin{tabular}{  p{\dimexpr 0.2\linewidth-2\tabcolsep} 
                   p{\dimexpr 0.1\linewidth-2\tabcolsep} 
                   p{\dimexpr 0.1\linewidth-2\tabcolsep} 
                   p{\dimexpr 0.1\linewidth-2\tabcolsep} 
                   p{\dimexpr 0.1\linewidth-2\tabcolsep}
                   p{\dimexpr 0.1\linewidth-2\tabcolsep} 
                   p{\dimexpr 0.1\linewidth-2\tabcolsep} 
                   p{\dimexpr 0.1\linewidth-2\tabcolsep} 
                   p{\dimexpr 0.1\linewidth-2\tabcolsep}                   
                   }
\toprule
\multirow{2}{*}{Dataset}    & \multirow{2}{*}{Encoder}         & \multirow{2}{*}{Version} & Linear       &       &        & MLP       &       &       \\ \cmidrule(l){4-9} 
        &  &      & $\%\Delta\textsubscript{Mem-Ex}$    & $\%\Delta\textsubscript{Mem-Freq}$    & $\%\Delta\textsubscript{Mem-Unif}$     & $\%\Delta\textsubscript{Mem-Ex}$    & $\%\Delta\textsubscript{Mem-Freq}$    & $\%\Delta\textsubscript{Mem-Unif}$   \\ \cmidrule(l){1-9}
\multirow{4}{*}{\conllchunking} & \multirow{2}{*}{\bert}  & base & 12.03 & 6.36  & 0.95   & 10.36 & 4.65  & 0.99  \\
        &  & random & \textbf{25.47} & \textbf{27.61} & \textit{0.44}   & \textbf{32.4}  & \textbf{34.6}  & \textbf{9.73}  \\ \cmidrule(l){2-9}
        & \multirow{2}{*}{\roberta} & base & 10.55 & 5.34  & 1.02   & 9.18  & 3.86  & 0.88  \\
        &  & random & \textbf{23.5}  & \textbf{26.36} & \textit{0.21}   & \textbf{31.69} & \textbf{34.45} & \textbf{10.65} \\ \cmidrule(l){1-9}
\multirow{4}{*}{\ontonotesconst} & \multirow{2}{*}{\bert}    & base & 13.34 & 4.38  & 6.91   & 10.75 & 3.33  & 5.62  \\
        &  & random & \textbf{17.42} & \textbf{7.66}  & \textbf{7.36}   & \textbf{18.91} & \textbf{7.02}  & \textbf{8.95}  \\
        & \multirow{2}{*}{\roberta} & base & 14.08 & 4.34  & 7.32   & 10.77 & 3.27  & 5.69  \\
        &  & random & \textbf{18.55} & \textbf{8.04}  & \textbf{7.97}   & \textbf{18.44} & \textbf{6.94}  & \textbf{8.68}  \\ \cmidrule(l){1-9}
\multirow{4}{*}{\ontonotessrl}   & \multirow{2}{*}{\bert}    & base & 7.47  & 1.9  & 6.27   & 5.15  & 0.87  & 4.27  \\
        &  & random & \textbf{12.77} & \textbf{2.73}  & \textbf{10.36}  & \textbf{14.63} & \textbf{3.01}  & \textbf{9.92} \\
        & \multirow{2}{*}{\roberta} & base & 7.83  & 1.25  & 6.52   & 5.12  & 0.83  & 4.29  \\
        &  & random & \textbf{12.7}  & \textbf{2.49}  & \textbf{10.37}  & \textbf{13.83} & \textbf{2.99}  & \textbf{9.21} \\ \cmidrule(l){1-9}
\multirow{4}{*}{\semeval}        & \multirow{2}{*}{\bert}    & base & 1.9  & 0.09  & 1.04   & 0.72  & 0.05  & 0.67  \\
        &  & random & \textit{0.75}  & \textit{-0.13} & \textit{0.84}   & \textit{0.19}  & \textit{0.04}  & \textit{0.27}  \\
        & \multirow{2}{*}{\roberta} & base & 1.4   & 0.04  & 1.3    & 0.78  & 0.02  & 0.72  \\
        &  & random & \textit{0.53}  & \textit{0}     & \textit{0.33}   & \textit{1.36}  & \textit{-0.04} & \textit{1.17}  \\ \bottomrule
\end{tabular}
\caption{The effect of heuristic algorithms on EP tasks where each task has \textit{only one} dataset. Each model is tested with the original test data and three \textit{filtered} test datasets. The $\%\Delta\textsubscript{Mem-Ex}$ shows the \textit{percentage drop in the accuracy score} from the original test dataset when the models are tested on the dataset filtered by ``Mem-Exact'' (the others follow the same nomenclature). \textbf{Bold} (\textit{Italicized}) indicates that the random encoder shows a much higher (lower) $\%$ drop on the filtered dataset than the base encoder.}
\label{tab:filtered-percent-drop-1-dataset}
\end{table*}

\begin{table*}[!hbt]
\footnotesize
\begin{tabular}{  p{\dimexpr 0.2\linewidth-2\tabcolsep} 
                   p{\dimexpr 0.1\linewidth-2\tabcolsep} 
                   p{\dimexpr 0.1\linewidth-2\tabcolsep} 
                   p{\dimexpr 0.1\linewidth-2\tabcolsep} 
                   p{\dimexpr 0.1\linewidth-2\tabcolsep}
                   p{\dimexpr 0.1\linewidth-2\tabcolsep} 
                   p{\dimexpr 0.1\linewidth-2\tabcolsep} 
                   p{\dimexpr 0.1\linewidth-2\tabcolsep} 
                   p{\dimexpr 0.1\linewidth-2\tabcolsep}                   
                   }
\toprule
\multirow{2}{*}{Dataset}    & \multirow{2}{*}{Encoder}         & \multirow{2}{*}{Version} & Linear       &       &        & MLP       &       &       \\ \cmidrule(l){4-9} 
        &  &      & $\%\Delta\textsubscript{Mem-Ex}$    & $\%\Delta\textsubscript{Mem-Freq}$    & $\%\Delta\textsubscript{Mem-Unif}$     & $\%\Delta\textsubscript{Mem-Ex}$    & $\%\Delta\textsubscript{Mem-Freq}$    & $\%\Delta\textsubscript{Mem-Unif}$   \\ \cmidrule(l){1-9}
\multirow{4}{*}{\ewtpos} & \multirow{2}{*}{\bert}    & base & 15.51 & 2.45  & 1.5    & 15.76 & 1.95  & 1.75  \\
        &  & random & \textbf{59.15} & \textbf{17.68} & \textit{-2.9}   & \textbf{57.34} & \textbf{15.67} & \textit{0.69}  \\ 
        & \multirow{2}{*}{\roberta} & base & 12.35 & 1.95  & 1.46   & 12.39 & 1.67  & 1.38  \\
        &  & random & \textbf{60.4}  & \textbf{17.54} & \textit{-2.88}  & \textbf{60.01} & \textbf{16.1}  & \textit{0}     \\ \cmidrule(l){2-9} 
\multirow{4}{*}{\ptbpos} & \multirow{2}{*}{\bert}    & base & 13.88 & 0.51  & 1.26   & 13.17 & 0.59  & 1.18  \\
        &  & random & \textbf{62.46} & \textbf{9.68}  & \textit{-2.86}  & \textbf{41.06} & \textbf{5.84}  & \textbf{1.92}  \\ 
        & \multirow{2}{*}{\roberta} & base & 13.94 & 0.66  & 1.06   & 12.89 & 0.54  & 1.17  \\
        &  & random & \textbf{69.64} & \textbf{10.63} & \textit{-3.71}  & \textbf{41.34} & \textbf{5.98}  & \textbf{2.03}  \\ \cmidrule(l){2-9}
\multirow{4}{*}{\ontonotespos}   & \multirow{2}{*}{\bert}    & base & 15.45 & 2.77  & 0.13   & 14.75 & 2.25  & 0.3   \\
        &  & random & \textbf{71.91} & \textbf{38.31} & \textit{-9.48}  & \textbf{65}    & \textbf{24.5}  & \textit{-2.94} \\ 
        & \multirow{2}{*}{\roberta} & base & 14.9 & 2.55  & 0.21   & 13.63 & 2.45  & 0.15  \\
        &  & random & \textbf{71.73} & \textbf{40.59} & \textit{-10.35} & \textbf{47.75} & \textbf{27.98} & \textit{-3.67} \\ \cmidrule(l){1-9}
\multirow{4}{*}{\conllner}       & \multirow{2}{*}{\bert}    & base & 9.16  & 0.63  & 3.56   & 10.67 & 0.6   & 4.07  \\
        &  & random & \textbf{34.47} & \textbf{2.43}  & \textbf{13.12}  & \textbf{32.56} & \textbf{2.81}  & \textbf{9.54} \\ 
        & \multirow{2}{*}{\roberta} & base & 8.62  & 0.58  & 3.47   & 8.23  & 0.48  & 3.19  \\
        &  & random & \textbf{34.1}  & \textbf{2.39}  & \textbf{12.98}  & \textbf{31.36} & \textbf{2.38}  & \textbf{9.7}  \\ \cmidrule(l){2-9}
\multirow{4}{*}{\ontonotesner}   & \multirow{2}{*}{\bert}    & base & 5.35 & 0.32 & 3.94   & 5.35  & -0.51 & 4.43  \\
        &  & random & \textbf{29.3}  & \textbf{14.98} & \textbf{5.15}   & \textbf{35.47} & \textbf{13.31} & \textbf{8.37}  \\ 
        & \multirow{2}{*}{\roberta} & base & 5.41  & 0.52  & 3.65   & 4.55  & -1.04 & 4.43  \\
        &  & random & \textbf{29.24} & \textbf{14.57} & \textbf{5.28}   & \textbf{29.26} & \textbf{9.43} & \textbf{7.03}  \\ \cmidrule(l){1-9}
\multirow{4}{*}{\ewtsyndepcls}   & \multirow{2}{*}{\bert}    & base & 8.7   & 0.13  & 8.36   & 6.72  & 0.46  & 6.09  \\
        &  & random & \textbf{29.36} & \textbf{0.69}  & \textbf{27.95}  & \textbf{26.98} & \textbf{2.01}  & \textbf{24.13} \\ 
        & \multirow{2}{*}{\roberta} & base & 8.12  & -0.04 & 7.91   & 6.74  & 0.48  & 6.04  \\ 
        &  & random & \textbf{30.48} & \textbf{0.86}  & \textbf{28.57}  & \textbf{26.94} & \textbf{1.88}  & \textbf{24.39} \\ \cmidrule(l){2-9}
\multirow{4}{*}{\ptbsyndepcls}   & \multirow{2}{*}{\bert}    & base & 9.23  & 0.14  & 7.92   & 6.76  & 0.48  & 5.2   \\
        &  & random & \textbf{36.12} & \textbf{0.97}  & \textbf{29}     & \textbf{32.28} & \textbf{2.35}  & \textbf{23.96} \\ 
        & \multirow{2}{*}{\roberta} & base & 9.44  & 0.29  & 8.03   & 6.72  & 0.51  & 5.19  \\
        &  & random & \textbf{36.76} & \textbf{0.73}  & \textbf{29.8}   & \textbf{32.17} & \textbf{2.25}  & \textbf{24.05} \\ \cmidrule(l){1-9}
\multirow{4}{*}{\ewtsyndeppred}  & \multirow{2}{*}{\bert}    & base & 4.52  & 0     & 4.59   & 5.27  & 1.29  & 4.14  \\
        &  & random & \textbf{5.66}  & \textbf{0.74}  & \textbf{4.95}   & \textit{4.71}  & \textit{1.14}  & \textit{3.46}  \\ 
        & \multirow{2}{*}{\roberta} & base & 6.64  & 0.91  & 5.57   & 5.05  & 1.18  & 3.97  \\
        &  & random & \textit{5.58}  & \textit{0.86}  & \textit{4.66}   & \textit{4.95}  & \textit{1.04}  & \textit{3.66}  \\ \cmidrule(l){2-9}
\multirow{4}{*}{\ptbsyndeppred}  & \multirow{2}{*}{\bert}    & base & 6.49  & 0.45  & 5.43   & 3.72  & 1.51  & 2.5   \\
        &  & random & \textit{4.77}  & \textit{0.3}   & \textit{3.68}   & \textit{2.67}  & \textit{1.17}  & \textit{1.93}  \\ 
        & \multirow{2}{*}{\roberta} & base & 7.47  & 0.96  & 5.73   & 4.58  & 2     & 2.91  \\
        &  & random & \textit{3.57}  & \textit{-0.09} & \textit{3.12}   & \textit{2.55}  & \textit{1.04}  & \textit{1.69}  \\ \cmidrule(l){1-9}
\multirow{4}{*}{\sprone} & \multirow{2}{*}{\bert}    & base & 0.38  & 0.04  & 0.31   & 0.35  & 0.05  & 0.29  \\
        &  & random & \textit{0.08}  & \textit{0}     & \textit{0.1}    & \textbf{0.66}  & \textit{-0.02} & \textbf{0.73}  \\ 
        & \multirow{2}{*}{\roberta} & base & 0     & 0     & -0.01  & 0.39  & 0.04  & 0.33  \\ 
        &  & random  & \textit{0.07}  & 0     & \textbf{0.08}   & \textit{0.36}  & \textbf{0.05}  & \textbf{0.36}  \\ \cmidrule(l){2-9}
\multirow{4}{*}{\sprtwo} & \multirow{2}{*}{\bert}    & base & 1.96  & 0     & 0.31   & 1.08  & 0     & 0.22  \\
        &  & random & \textit{1.37}  & 0     & \textit{0.17}   & \textbf{1.8}   & 0     & \textbf{0.29}  \\
        & \multirow{2}{*}{\roberta} & base & 1.55  & 0     & 0.28   & 1.65  & 0     & 0.34  \\
        &  & random & 1.55  & 0     & \textit{0.24}   & \textit{1.5}   & 0     & \textit{0.26}  \\ \cmidrule(l){1-9}
\multirow{4}{*}{\dprd}   & \multirow{2}{*}{\bert}    & base & 4.37  & 0     & 1.52   & 0.73  & -0.22 & 1.13  \\
        &  & random & \textit{0.92}  & \textit{-0.2}  & \textit{2.91}   & \textit{0.92}  & \textit{0.22}  & \textit{0.2}   \\
        & \multirow{2}{*}{\roberta} & base & 1.26  & 0.2   & 1.93   & 0.81  & 0.19  & 3.34  \\
        &  & random & \textit{0.38}  & \textit{0.22}  & \textit{0.84}   & \textit{-0.5}  & \textit{-0.22} & \textit{-1.26} \\ \cmidrule(l){2-9}
 \multirow{4}{*}{\ontonotescoref}   & \multirow{2}{*}{\bert}    & base & -2.92  & -1.61     & -0.99   & 0.84  & 0.58 & 0.88  \\
        &  & random & \textit{-7.42}  & \textit{-4.59}  & \textit{-3.18}   & \textbf{0.78}  & \textit{1.55}  & \textit{0.72}   \\
        & \multirow{2}{*}{\roberta} & base & -3.47  & -1.66   & -1.45   & 1.05  & 0.87  & 0.86  \\
        &  & random & \textit{-6.76}  & \textit{-4.84}  & \textit{-2.46}   & \textit{0.4}  & \textbf{1.15} & \textit{0.58} \\   
\bottomrule
\end{tabular}
\caption{The effect of heuristic algorithms on EP tasks where each task has \textit{multiple} datasets. The structure follows \cref{tab:filtered-percent-drop-1-dataset}.}
\label{tab:filtered-percent-drop-2-dataset}
\end{table*}

For each heuristic algorithm in \cref{sec:rq1}, we create a ``filtered dataset''  consisting of the points that can not be classified using the said algorithm. For each ``EP model'' (an encoder + EP classifier), we calculate the accuracy score on the original and the filtered datasets and report the ``drop'', i.e., the relative reduction percentage:  $(acc\textsubscript{original} - acc\textsubscript{filtered})*100/acc\textsubscript{original})$. A \textit{negative} drop indicates that the EP model performed \textit{better} on the original dataset vs the filtered one.

Tables~\ref{tab:filtered-percent-drop-1-dataset} and ~\ref{tab:filtered-percent-drop-2-dataset} show the results. Firstly, there is an accuracy drop in both pre-trained (base) and random encoders with \textbf{all} ``Mem-Exact'' datasets, indicating these datasets are more difficult in general and both these encoders use the exact memorization heuristic \cite{DBLP:journals/csl/AugensteinDB17} to some extent. On the other hand, they do not use the baseline ``Mem-Uniform'' heuristic as expected, as evidenced by the increased accuracy in the filtered dataset. 

More importantly, in a large number of EP datasets (11 out of 17), the accuracy drop in the random encoder is higher (indicated by \textbf{bold}) than that in the pre-trained encoders. Also, this pre-trained-v-random accuracy drop difference in the filtered datasets is \textbf{significant}, i.e., $> 100\%$, in 8 out of 11 cases. On the other hand, when the random encoders show a lower drop than the pre-trained encoders, the difference is almost always negligible (eg., \ewtsyndeppred). In 4 of the remaining 6 datasets where we do not see a higher drop in the random encoders - \semeval, \sprone, \sprtwo, and \dpr, the filtered version of the datasets do not differ much from the original: as only a small percentage of the data points can be solved by the \textbf{Mem-Exact} heuristic.

The accuracy drops are consistent across the encoder types and EP classifiers. For example, on the \ewtpos\ dataset, the \bert-base and the \roberta-base encoders have similar drops both with the Linear and the MLP EP classifiers as do the random versions of these encoders among themselves. A surprising finding is that the drop pattern is task-dependent. Among the tasks with multiple datasets (\cref{tab:filtered-percent-drop-2-dataset}), in all POS, NER, and Syntactic Dependency Classification datasets, the random encoders show a higher drop but in the Syntactic Dependency Prediction and \spr\ tasks, the opposite is true for all datasets. This is not correlated with either the dataset size or the number of labels: both Syntactic Dependency Prediction and Classification tasks have a similar number of training data points, and the Classification task has $\approx$ 40 labels whereas the Prediction one has only 2. 

The \ontonotescoref\ dataset presents an interesting case as the accuracy scores \textbf{increase} in the filtered datasets. This binary classification dataset has a significant label imbalance: 78.33\% of the test data has a negative label. If the dataset is re-sampled to make the distribution balanced, a) the accuracy score decreases as expected; b) the accuracy drops in the random encoders become higher by 19.28 and 7.08 points than the \bert\ and \roberta\ encoders respectively when using the MLP classifier. With the Linear classifier, these numbers are 3.45 and 8.7.     

Overall, it is clear that in many EP test datasets, the random encoders perform significantly worse than the pre-trained encoders on the set of data points that are \textbf{not} heuristically classifiable (specifically, by the \textbf{Mem-Exact} heuristic). In other words, they resort to the heuristics more than the pre-trained ones. This proves our hypothesis.  

\section{EP Test Results: Random vs Pre-Trained Encoders}
\label{sec:rq3}

Previously, we have shown that the random encoders show a significant memorization bias compared to the pre-trained ones. How does that affect the EP test results? \cref{tab:acc-base-v-random} and \cref{tab:acc-base-v-random-mlp} show the EP test results for the pre-trained and random encoders on the ``Mem-Exact'' filtered datasets - except for the \ontonotescoref\ one, where we use the \textit{balanced} dataset. As expected, in almost all cases the pre-trained encoders have a \textbf{significantly higher} accuracy than the random ones. Compare this with \citet{DBLP:conf/emnlp/VoitaT20} where in 4 out of 7 datasets that is not the case.

\begin{table}[!hbt]
\footnotesize
\centering
\begin{tabular}{  p{\dimexpr 0.4\linewidth-2\tabcolsep} 
                   p{\dimexpr 0.15\linewidth-2\tabcolsep} 
                   p{\dimexpr 0.15\linewidth-2\tabcolsep} 
                   p{\dimexpr 0.15\linewidth-2\tabcolsep} 
                   p{\dimexpr 0.15\linewidth-2\tabcolsep} }
\toprule
\multirow{2}{*}{Dataset} & \multicolumn{2}{l}{\bert} & \multicolumn{2}{l}{\roberta} \\ 
& pre-trained & random & pre-trained & random \\ \cmidrule(l){1-5}
\ewtpos & 79.74 & \textbf{25.93} & 83.71 & \textbf{24.61} \\
\ptbpos & 83.33 & \textbf{24.79} & 83.52 & \textbf{19.90} \\ 
\ontonotespos & 81.62 & \textbf{17.43} & 83.30 & \textbf{17.29} \\ \cmidrule(l){1-5}
\conllner & 87.96 & \textbf{54.26} & 88.65 & \textbf{54.42} \\ 
\ontonotesner & 87.95 & \textbf{35.83} & 88.83 & \textbf{35.11} \\ \cmidrule(l){1-5}
DPR & 48.37 & \textit{49.70} & 50.15 & 49.55 \\
\ontonotescoref & 70.53 & \textbf{60.41} & 72.9 & \textbf{58.44} \\ \cmidrule(l){1-5}
\ewtsyndepcls & 69.35 & \textbf{32.60} & 71.78 & \textbf{31.36} \\
\ptbsyndepcls & 78.58 & \textbf{33.59} & 79.19 & \textbf{32.96} \\ \cmidrule(l){1-5}
\ewtsyndeppred & 66.54 & \textbf{62.66} & 67.72 & \textbf{63.63} \\
\ptbsyndeppred & 64.45 & 63.14 & 63.93 & \textit{64.20} \\ \cmidrule(l){1-5}
\sprone$^*$ & 70.68 & \textbf{60.66} & 67.21 & \textbf{61.38} \\
\sprtwo$^*$ & 75.12 & \textbf{69.88} & 76.61 & \textbf{70.41} \\ \cmidrule(l){1-5}
\conllchunking & 81.43 & \textbf{50.29} & 84.40 & \textbf{50.81} \\
\ontonotesconst & 62.17 & \textbf{38.15} & 62.81 & \textbf{38.21} \\
\ontonotessrl & 67.79 & \textbf{44.45} & 68.71 & \textbf{44.96} \\
\semeval & 55.10 & \textbf{22.39} & 50.88 & \textbf{24.35} \\ 
\bottomrule
\end{tabular}
\caption{Accuracy scores (Micro f1 for $^*$) on the filtered EP test dataset, with the \textbf{Linear} classifier. \textbf{Bold} indicates where the random encoders have a \textit{significantly lower} score than the pre-trained ones, and \textit{Italicized} indicates they have a higher score.}
\label{tab:acc-base-v-random}
\end{table}

\begin{table}[!hbt]
\footnotesize
\centering
\begin{tabular}{  p{\dimexpr 0.4\linewidth-2\tabcolsep} 
                   p{\dimexpr 0.15\linewidth-2\tabcolsep} 
                   p{\dimexpr 0.15\linewidth-2\tabcolsep} 
                   p{\dimexpr 0.15\linewidth-2\tabcolsep} 
                   p{\dimexpr 0.15\linewidth-2\tabcolsep} }
\toprule
\multirow{2}{*}{Dataset} & \multicolumn{2}{l}{\bert} & \multicolumn{2}{l}{\roberta} \\ 
& pre-trained & random & pre-trained & random \\ \cmidrule(l){1-5}
\ewtpos & 79.93 & \textbf{31.48} & 84.33 & \textbf{28.83} \\
\ptbpos & 84.31 & \textbf{48.84} & 84.86 & \textbf{47.99} \\ 
\ontonotespos & 82.98 & \textbf{27.98} & 84.17 & \textbf{40.82} \\ \cmidrule(l){1-5}
\conllner & 86.6 & \textbf{57.34} & 89.44 & \textbf{58.14} \\ 
\ontonotesner & 84.55 & \textbf{38.53} & 87.38 & \textbf{41.77} \\ \cmidrule(l){1-5}
DPR & 59.94 & \textbf{49.7} & 51.63 & 50.3 \\
\ontonotescoref & 85.91 & \textbf{73.09} & 87.4 & \textbf{73.12} \\ \cmidrule(l){1-5}
\ewtsyndepcls & 80.57 & \textbf{42.43} & 81.63 & \textbf{42.35} \\
\ptbsyndepcls & 86.85 & \textbf{44.12} & 87.42 & \textbf{44} \\ \cmidrule(l){1-5}
\ewtsyndeppred & 79.26 & \textbf{72.65} & 81.38 & \textbf{73.41} \\
\ptbsyndeppred & 86.72 & \textbf{80.05} & 86.19 & \textbf{81.17} \\ \cmidrule(l){1-5}
\sprone$^*$ & 81.97 & \textbf{63.68} & 83.7 & \textbf{63.5} \\
\sprtwo$^*$ & 77.91 & \textbf{72.06} & 77.31 & \textbf{71.5} \\ \cmidrule(l){1-5}
\conllchunking & 84.88 & \textbf{50.15} & 86.97 & \textbf{50.54} \\
\ontonotesconst & 70.55 & \textbf{49} & 71.05 & \textbf{49.44} \\
\ontonotessrl & 80.26 & \textbf{51.06} & 80.86 & \textbf{51.34} \\
\semeval & 65.04 & \textbf{26.01} & 63.8 & \textbf{26.03} \\ 
\bottomrule
\end{tabular}
\caption{Accuracy scores (Micro f1 for $^*$) on the filtered EP test dataset, random vs pre-trained encoders with the \textbf{MLP} classifier. \textbf{Bold} indicates where the random encoders have a \textit{significantly lower} score than the pre-trained ones.}
\label{tab:acc-base-v-random-mlp}
\end{table}

\textbf{MDL Probe.} 
\citet{DBLP:conf/emnlp/VoitaT20} show that for many EP datasets, a contextual encoder (\texttt{ElMo}) has the same performance as a random encoder. This leads to the conclusion that the EP tests, in reality, measure the classifiers' ability to learn the EP task and do not reflect the knowledge encoded in the representations themselves. To solve this, a minimum description length (MDL) probe is proposed. We have already seen that the pre-trained vs random issue is mitigated in the filtered datasets, but had we used the MDL probes, would our conclusions have changed? More importantly, are the MDL probes necessary in the EP test datasets with a large number of samples (\cref{table:dataset})?  

In its original formulation, the Minimum Description Length (MDL) principle is a Bayesian model selection technique. A model class $\mathbf{M}$ is a set of models $M_i$, for example, $\mathbf{M}$ can be ``all polynomials of degree 3'' and one $M_i$ can be $5x^3$. Between two model classes $\mathbf{M_a}$ and $\mathbf{M_b}$, the better model class is the one with the lower \textit{stochastic complexity}. 

Given a supervised classification dataset $D$ with data points $d_i = \langle x_i, y_i \rangle$, a model $M$ defines a probability distribution $P(y_i|x_i)$. From the Kraft-Mcmillan inequality, there exists a code $C$ for $D$ with the code length $L_C(D) = -logP(D) = \sum_{i=1}^{n}-logP(d_i)$. Naturally, a better model fit corresponds to higher probability values and lower code lengths.

The stochastic complexity of the dataset $D$ with respect to the model class $\mathbf{M}$ is the shortest code length of $D$ when $D$ is encoded with the help of class $\mathbf{M}$. Given $M$ and $D$, one can find the $M_i$ (with parameters $\theta_i$) through maximum likelihood estimation that leads to the maximum $P$, hence the minimum code length $L(D|\hat{\theta}(D)) = -logP(D|\hat{\theta}(D))$.

Crucially, we are not allowed to fit a different $\theta$ and build a new code $C'$ with each new dataset $D'$. Ideally, we would like to have a single code $C^{*}$ that can yield the minimum length for \textit{all} datasets but that is not possible if $\mathbf{M}$ contains more than one model. Nevertheless, it is possible to construct $C^{*}$ such that: \cite{grunwald2000model}
\begin{gather}
\label{eq:3}
L_{C^{*}}(D) = L(D|\hat{\theta}(D)) + K^{*}
\end{gather}
Equation~\ref{eq:3} is a combination of the ``goodness of model fit'' (better estimate of $\hat{\theta}$ $\implies$ smaller code length) and the model complexity ($K^{*}$). $K^{*}$ can be approximated for a \textit{regular} model class $\mathbf{M}$ containing models with $p$ parameters as: 

\begin{gather}
\label{eq:4}
K^{*} \approx \frac{p}{2}log n + C_k
\end{gather}
where $n$ is the length of the dataset $D$ and $C_k$ is negligible for large $n$ \cite{grunwald2000model}. 

\citet{DBLP:conf/emnlp/VoitaT20} calculate the code lengths of two EP classifiers with random and pre-trained encoders and show that the second one has a lower code length. This is one of the reasons for using the minimization of codelengths (which is termed ``MDL probe'') as an alternative to normal classifiers. In the implementation, these two encoders are frozen and hence provide two \textit{datasets}, so the model selection problem is essentially inverted: there is one model class (say, the class of Linear models) and two datasets (token encodings from random and pre-trained encoders): what would two different code lengths mean?

\citet{DBLP:conf/emnlp/VoitaT20} follows \citet{DBLP:conf/nips/BlierO18} in determining code lengths for DNN models because the approximation in \cref{eq:4} is not correct for complex DNNs. But the EP classifiers are not DNNs, they are simple linear models whose code lengths should be approximable by \cref{eq:4}. But as \cref{eq:4} shows, the code lengths are not dependent on the datasets as long as the number of data points is large, which is true for most EP datasets (see \autoref{table:dataset}). This raises the question of whether the MDL probe is an inherently better choice for comparing the encoding of information in the encoders. 

\section{Related Work}
\label{sec:related-work}
Previous research has primarily focused on studying different aspects of pre-trained language models (LMs), such as linguistic knowledge \cite{liu-etal-2019-linguistic} and attention patterns \cite{clark-etal-2019-bert}. 

The paradigm of classifier-based probing tasks is well-researched \cite{DBLP:conf/repeval/EttingerER16} and has gained popularity with the introduction of benchmark EP datasets that we utilize here \cite{tenney-etal-2019-bert}. Typically, internal layers of large language or machine translation models are used as features for auxiliary prediction tasks related to syntactic properties, such as part-of-speech \cite{DBLP:conf/emnlp/ShiPK16, DBLP:conf/acl/BlevinsLZ18, DBLP:conf/acl/TenneyDP19}, tense \cite{DBLP:conf/emnlp/ShiPK16, DBLP:conf/acl/TenneyDP19}, or subject-verb agreement \cite{DBLP:conf/emnlp/TranBM18, DBLP:journals/tacl/LinzenDG16}. For a comprehensive survey, refer to \citet{DBLP:journals/tacl/BelinkovG19}.

EP tests are not direct evaluations of models since they use another model (called probe) to extract and evaluate the linguistic features within an encoding. Because of this, it is not clear if the results reflect the quality of encoding or the probe's ability to learn the task \cite{DBLP:conf/emnlp/HewittL19, DBLP:conf/emnlp/VoitaT20, pimentel-etal-2020-information}.
We delve into this topic further in Section \ref{sec:rq3}. Additional details can be found in \citet{belinkov-2022-probing}.

\section{Conclusion}
EP tests are classification tasks to measure an LLM's ability to encode syntactic and semantic knowledge. However, in many EP datasets, there is not a significant difference between the random vs pre-trained encoders, which raises questions about the validity of the tests (the ``classifier knowledge'' problem). We analyze 17 datasets across 10 datasets to find various biases and show that the EP classifiers are more prone to use heuristic mechanisms when random encoders are used instead of the pre-trained ones. When the dataset biases are removed, the pre-trained encoders do show a significant difference from the random ones as expected. Information-theoretic probes have been proposed before to solve the ``classifier knowledge'' problem, we show why they might not be necessary. Future work would extend the findings of this study to fine-tuned models.

\section*{Limitations}
There are two important limitations of this study: 1. We analyze a large number of standardized EP test datasets that have been extensively used before, but the paradigm of diagnostic classifiers is quite broad and our findings should not be automatically extended to datasets not used in this study. Also, we do not propose an automated way to remove biases from the existing or newly created datasets. 2. While we argue the popular MDL probe might not be necessary for all EP test datasets (particularly, the ones with a large number of data points), this paper should not be construed as a general criticism of the MDL probes or the area of information-theoretic probing.    

\section*{Acknowledgement}

We thank the anonymous reviewers for their helpful suggestions.

\bibliography{acl}
\bibliographystyle{acl_natbib}
\clearpage


\end{document}